\documentclass{article}

\usepackage{microtype}
\usepackage{graphicx}
\usepackage{subcaption}
\usepackage{booktabs} % for professional tables

\usepackage{hyperref}

\usepackage[preprint]{icml2026}

\usepackage{amsmath}
\usepackage{amssymb}
\usepackage{mathtools}
\usepackage{amsthm}

\usepackage[capitalize,noabbrev]{cleveref}

\theoremstyle{plain}

\theoremstyle{definition}

\theoremstyle{remark}

\usepackage[T1]{fontenc}

\usepackage[textsize=tiny]{todonotes}

\icmltitlerunning{Spherical Latent Motion Prior for Physics-Based Simulated Humanoid Control}

\begin{document}

\twocolumn[
  \icmltitle{Spherical Latent Motion Prior for Physics-Based Simulated Humanoid Control}

  % \vspace{0.5em}
   
    % \vspace{0.5em}

  % It is OKAY to include author information, even for blind submissions: the
  % style file will automatically remove it for you unless you've provided
  % the [accepted] option to the icml2026 package.

  % List of affiliations: The first argument should be a (short) identifier you
  % will use later to specify author affiliations Academic affiliations
  % should list Department, University, City, Region, Country Industry
  % affiliations should list Company, City, Region, Country

  % You can specify symbols, otherwise they are numbered in order. Ideally, you
  % should not use this facility. Affiliations will be numbered in order of
  % appearance and this is the preferred way.
  \icmlsetsymbol{equal}{*}
  \icmlsetsymbol{Corresponding author}{†}

    \begin{icmlauthorlist}
        \icmlauthor{Jing Tan}{equal,yyy}
        \icmlauthor{Weisheng Xu}{equal,yyy}
        \icmlauthor{Xiangrui Jiang}{equal,yyy}
        \icmlauthor{Jiaxi Zhang}{yyy}
        \icmlauthor{Kun Yang}{yyy}
        \icmlauthor{Kai Wu}{yyy}
        \icmlauthor{Jiaqi Xiong}{oxf}
        \icmlauthor{Shiting Chen}{yyy}
        \icmlauthor{Yangfan Li}{yyy}
        \icmlauthor{Yixiao Feng}{yyy}
        \icmlauthor{Yuetong Fang}{yyy}
        \icmlauthor{Yujia Zou}{yyy}
        \icmlauthor{Yiqun Song}{yyy}
        \icmlauthor{Renjing Xu}{Corresponding author,yyy}
    \end{icmlauthorlist}
    
    \icmlaffiliation{yyy}{The Hong Kong University of Science and Technology (Guangzhou), Guangzhou, China}
    \icmlaffiliation{oxf}{University of Oxford, Oxford, United Kingdom}
    
    \icmlcorrespondingauthor{Renjing Xu}{renjingxu@hkust-gz.edu.cn}

    \begin{center}
    % \small
    Project page: \href{https://colin-jing.github.io/SLMP/}{https://colin-jing.github.io/SLMP/}
    \end{center}
  % You may provide any keywords that you find helpful for describing your
  % paper; these are used to populate the "keywords" metadata in the PDF but
  % will not be shown in the document
  % \icmlkeywords{Machine Learning, ICML}

  \vskip 0.3in
]

% this must go after the closing bracket ] following \twocolumn[ ...

% This command actually creates the footnote in the first column listing the
% affiliations and the copyright notice. The command takes one argument, which
% is text to display at the start of the footnote. The \icmlEqualContribution
% command is standard text for equal contribution. Remove it (just {}) if you
% do not need this facility.

% Use ONE of the following lines. DO NOT remove the command.
% If you have no special notice, KEEP empty braces:
\printAffiliationsAndNotice{\icmlEqualContribution}  % no special notice (required even if empty)
% Or, if applicable, use the standard equal contribution text:
% \printAffiliationsAndNotice{\icmlEqualContribution}

\begin{abstract}
Learning motion priors for physics-based ss humanoid control is an active research topic. Existing approaches mainly include variational autoencoders (VAE) and adversarial motion priors (AMP). VAE introduces information loss, and random latent sampling may sometimes produce invalid behaviors. AMP suffers from mode collapse and struggles to capture diverse motion skills. We present the Spherical Latent Motion Prior (SLMP), a two-stage method for learning motion priors. In the first stage, we train a high-quality motion tracking controller. 
In the second stage, we distill the tracking controller into a spherical latent space. A combination of distillation, a discriminator, and a discriminator-guided local semantic consistency constraint shapes a structured latent action space, allowing stable random sampling without information loss.
To evaluate SLMP, we collect a two-hour human combat motion capture dataset and show that SLMP preserves fine motion detail without information loss, and random sampling yields semantically valid and stable behaviors. When applied to a two-agent physics-based combat task, SLMP produces human-like and physically plausible combat behaviors only using simple rule-based rewards. Furthermore, SLMP generalizes across different humanoid robot morphologies, demonstrating its transferability beyond a single simulated avatar.
\end{abstract}

\section{Introduction}

Physics-based simulated humanoid control has made notable progress in recent years, enabling simulated humanoids to imitate large-scale motion capture datasets, perform agile movements \cite{peng2018deepmimic,luo2021dynamics,luo2023perpetual,tessler2024maskedmimic, zhang2025ADD, jing2025farm}, and interact with the environment \cite{gao2024coohoi, pan2025tokenhsi, wang2025skillmimic}.
 These advances benefit applications in graphics, gaming, virtual reality, and robotics, where physically plausible human motion is essential. Despite these gains, controlling high-dimensional humanoids through raw joint torques remains challenging due to unstable dynamics and the sensitivity of low-level control to reward design \cite{luo2024universal}. To mitigate these issues, many recent methods introduce motion priors that constrain actions to physically plausible and coordinated motion manifolds, thereby simplifying policy learning for downstream tasks \cite{peng2022ase, tessler2023calm, luo2024universal, luo2024smplolympics, tessler2024maskedmimic}.

A key challenge in learning motion priors is achieving both diversity and validity. VAE–based \cite{kingma2013auto}  approaches \cite{won2021control,zhu2023neural,luo2024universal} compress full-body motion trajectories into latent codes, but the bottleneck often discards important motor details, and random latent sampling can produce implausible behaviors or falls. 
% AMP-based approaches \cite{peng2022ase, tessler2023calm} improve realism by matching discriminator \cite{goodfellow2020generative} statistics to a motion dataset, 
% but they tend to suffer from mode collapse and provide limited semantic coverage, which makes them hard to scale to diverse motion collections.
AMP-based \cite{peng2021amp} approaches \cite{peng2022ase, tessler2023calm} incorporate adversarial reward shaping, where a discriminator \cite{goodfellow2020generative} distinguishes real from simulated motion and nudges the policy toward the motion data distribution. However, these methods often suffer from mode collapse and limited semantic coverage, making them difficult to scale to diverse motion datasets.
As a result, existing priors struggle to support broad motion repertoires, stable random sampling, and reliable hierarchical reuse.

These limitations arise from structural properties of the underlying formulations. VAE-based priors instead compress high-dimensional, multi-modal human motions through a Gaussian latent bottleneck, which encourages a unimodal latent distribution and blurs fine-grained variations, while leaving large regions of latent space with little training signal so that random sampling may produce implausible actions. AMP-based priors rely on reinforcement learning under adversarial rewards, which couples high-variance policy gradients with discriminator feedback that emphasizes a few high-reward behaviors. This often drives the policy toward a small set of modes rather than a well-covered motion repertoire. 
Since humanoid motion repertoires are inherently multi-modal and state-dependent, a suitable latent space should reflect meaningful local structure for control while avoiding excessive information compression and unbounded low-density regions that destabilize random sampling.
% These observations suggest that a more suitable latent space for humanoid control should (i) avoid excessive information compression, (ii) avoid unbounded low-density tails that dominate random sampling, and (iii) provide well-structured latent neighborhoods that can represent both multi-modal and single-modal behaviors in a state-dependent manner.

Motivated by these requirements, we propose the Spherical Latent Motion Prior (SLMP), a two-stage method that constructs a structured latent action space for physics-based humanoids. In the first stage, we train a high-quality motion tracking controller. In the second stage, we distill this controller into a unit-sphere latent space by conditioning the policy on latent codes. The distillation objective combines three components: imitation distillation from the expert controller, a discriminator loss that distinguishes expert from non-expert actions, and a discriminator-guided local semantic consistency loss that shapes coherent neighborhoods on the sphere. This formulation preserves motor detail without a reconstruction bottleneck while enabling stable and diverse random sampling.

To evaluate SLMP, we collect a two-hour combat motion capture dataset spanning strikes, evasions, transitions, and footwork. Experiments show that SLMP preserves fine motion detail, exhibits stronger random sampling performance than VAE-based and AMP-based baselines, and produces state-dependent latent neighborhoods that encode feasible action sets. When used as a prior in a two-agent physics-based combat task, SLMP enables simple high-level policies to generate physically plausible and diverse behaviors using lightweight reward signals. Finally, we demonstrate that the same pipeline generalizes to real humanoid robot platforms. Our contributions are summarized as follows:
\begin{itemize}
    \item We collect and will release a two-hour human combat motion capture dataset, featuring diverse striking, evasion, and footwork behaviors suitable for physics-based humanoid control research.
    \item We propose SLMP, a two-stage distillation framework that constructs a structured spherical latent action space for humanoids, enabling stable and diverse latent-conditioned control.
    \item Through extensive experiments, we demonstrate that SLMP enables physically plausible two-agent combat using simple rule-based reward functions and generalizes across different humanoid robot morphologies.
\end{itemize}

\vspace{-5pt}

\section{Related Work}

\subsection{Physics-based Humanoid Motion tracking.}
Early physics-based systems such as DeepMimic \cite{peng2018deepmimic} demonstrated that reinforcement learning can track reference motion capture clips within a simulated environment. UHC \cite{luo2021dynamics} enabled general-purpose full-body tracking across AMASS \cite{mahmood2019amass} sequences, while PHC \cite{luo2023perpetual} removed external force modules and achieved more stable tracking on large-scale datasets. PHC+ \cite{luo2024universal} further refined training to obtain near-perfect tracking for everyday motions, and MaskedMimic \cite{tessler2024maskedmimic} introduced masked motion inpainting to support versatile conditioning modes in a single tracking model. Most recently, FARM \cite{jing2025farm} focused on high-dynamic motions and combined frame-accelerated augmentation with a residual mixture-of-experts architecture to better handle rapid pose transitions in physics-based tracking.

\subsection{Physics-based Humanoid Motion Latent Space}

Latent motion spaces for physics-based control mainly follow two design families. AMP-based methods such as ASE \cite{peng2022ase} learn adversarial skill embeddings on a spherical manifold for downstream control, while CALM \cite{tessler2023calm} learns conditional adversarial latent models for controllable character manipulation. These approaches support skill composition but often exhibit limited latent coverage and sampling stability on highly diverse datasets. 
VAE-based methods instead learn reconstruction-driven motion priors. PULSE \cite{luo2024universal} and MaskedMimic \cite{tessler2024maskedmimic} adopt VAE-style latent action models for character control, while Neural Categorical Priors \cite{zhu2023neural} use VQ-VAE-style \cite{van2017neural} discrete latents to encode behavioral modes. However, reconstruction-based objectives can introduce information bottlenecks and reduce sampling reliability.
In contrast, our method learns a continuous spherical latent space through controller distillation, focusing on stable and semantically meaningful sampling for physics-based humanoid control.

% \subsection{Physics-based Humanoid Motion Latent Space}
% Latent motion spaces for physics-based control mainly follow two design families. AMP-based methods such as ASE \cite{peng2022ase} learn adversarial skill embeddings on a spherical manifold for downstream control, while CALM \cite{tessler2023calm} learns conditional adversarial latent models that enable directable character manipulation. These approaches enable interactive skill switching but remain limited in latent coverage and sampling stability on large and highly diverse datasets. VAE-based methods instead learn reconstruction-driven motion priors: PULSE \cite{luo2024universal} and MaskedMimic \cite{tessler2024maskedmimic} adopt a VAE-style latent action model for character control. Neural Categorical Priors \cite{zhu2023neural} employ VQ-VAE-style \cite{van2017neural} discrete latent spaces to represent behavioral modes for downstream control. Compared to these, our approach avoids VAE bottlenecks. We distill a high-quality controller into a continuous spherical latent space using a combination of distillation, a discriminator for distinguishing expert from non-expert actions, and discriminator-guided semantic consistency, enabling stable random sampling.

\section{Method}
\label{sec:method}

\begin{figure*}[t]
    \centering
    % place your pipeline figure here
    \includegraphics[width=1.0\linewidth]{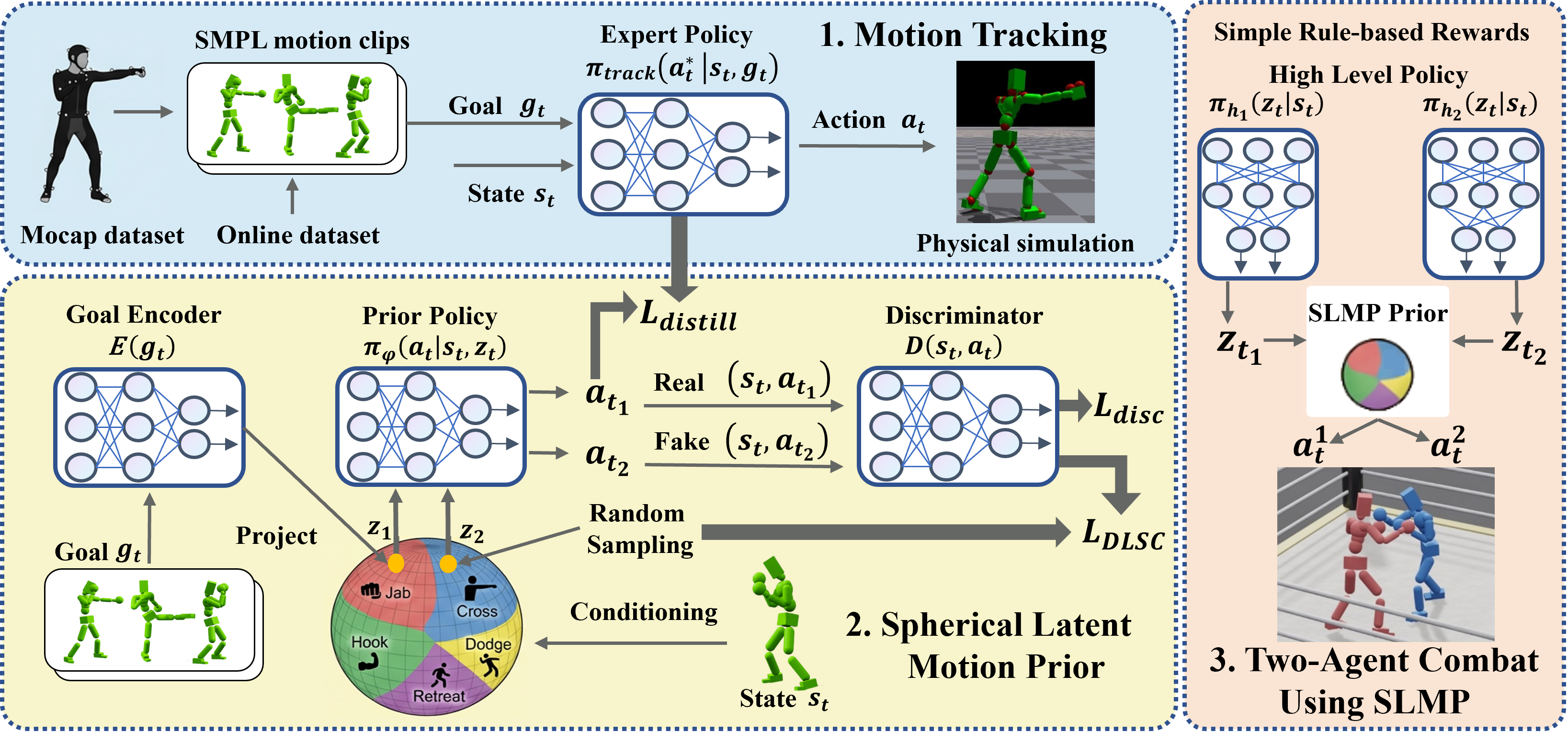}
    \caption{Overview of the Spherical Latent Motion Prior (SLMP). We collect a two-hour combat motion capture dataset and convert it into SMPL motion clips for training. We train a goal-conditioned motion tracking controller, then distill it into a unit-sphere latent space using three losses: imitation, discriminator, and our discriminator-guided local semantic consistency loss ($L_{\text{DLSC}}$). SLMP supports meaningful random sampling and drives downstream tasks such as two-agent combat via simple rewards.}
    \label{fig:pipeline_overview}
\end{figure*}

Our method follows the pipeline shown in Figure~\ref{fig:pipeline_overview}. We first collect a two-hour human combat mocap dataset and convert it into SMPL \cite{SMPL:2015} motion clips. Based on these clips, we train a goal-conditioned expert controller $\pi_{\text{track}}$ via motion tracking (Section~\ref{sec:pretrain}). We then distill this expert into a spherical latent space, using a discriminator and our discriminator-guided local semantic consistency loss to shape the latent manifold (Section~\ref{sec:slmp}). 
Finally, we use SLMP as a structured low-level prior in a two-agent combat task. A high-level policy outputs latent codes that drive the SLMP controller, while self-play \cite{silver2017mastering, berner2019dota} optimizes these latent decisions using simple rule-based rewards (Section~\ref{sec:high_level}).
% Finally, we use SLMP as a structured low-level prior in a two-agent combat task, where high-level policies operate in latent space and self-play discovers competitive behaviors under simple rule-based rewards (Section~\ref{sec:high_level}).

%--- 整体 Pipeline 图 ---

% \vspace{-1pt}

\subsection{Motion Tracking via Goal-Conditioned Reinforcement Learning}
\label{sec:pretrain}

We train a goal-conditioned motion tracking controller $\pi_{\text{track}}$ that maps the humanoid’s proprioceptive state and a reference-driven goal to per-joint target poses. This formulation follows the goal-conditioned reinforcement learning (GCRL) framework widely used in large-scale humanoid controllers. 
At each timestep $t$, the policy observes the humanoid state $s_t$ together with a goal $g_t$ derived from the SMPL reference motion. 
% which provides target joint rotations and linear and angular velocities for imitation.
% At each timestep $t$, the policy observes the humanoid state $s_t$ and a goal $g_t$ containing target joint orientations and velocities computed from the SMPL reference motion. 
The policy outputs PD target angles $a_t$ for each actuated joint, and joint torques are applied by the simulator through a PD controller. We optimize $\pi_{\text{track}}$ using PPO \cite{schulman2017proximal} to maximize a tracking reward that encourages physical imitation of the reference motion. The reward includes exponential matching terms for joint positions, joint orientations, linear velocities, and angular velocities, together with an energy penalty that discourages excessive torque expenditure. After training, $\pi_{\text{track}}$ achieves reliable tracking performance on our combat motion dataset and serves as the expert controller for our latent distillation stage.

% We build a base motion tracking controller $\pi_{\text{track}}$ that maps a physics state and an imitation goal to physically valid joint-space actions \cite{luo2023perpetual}. This module follows the goal-conditioned reinforcement learning (GCRL) framework \cite{schaul2015universal} widely adopted in universal humanoid controllers. At each timestep $t$, the policy receives the humanoid's proprioceptive state $s_t$, including joint positions and velocities of all SMPL joints, together with a reference-driven goal $g_t$ that contains target joint poses and velocities from the SMPL motion clips. The policy outputs PD targets $a_t$ for each actuated degree of freedom, and the simulator applies the corresponding torques. We optimize $\pi_{\text{track}}$ using PPO \cite{schulman2017proximal} to maximize a tracking reward that encourages physical imitation of the reference motion. The reward consists of several matching terms for joint positions, joint rotations, linear velocities, and angular velocities, plus an energy term that penalizes excessive torque usage. After training, $\pi_{\text{track}}$ reliably tracks motions from our combat dataset and provides the expert demonstrations needed for the latent distillation stage.

\subsection{Spherical Latent Motion Prior (SLMP)}
\label{sec:slmp}

We learn a latent-conditioned motion prior that distills $\pi_{\text{track}}$ into a spherical latent space. The training procedure of SLMP is summarized in Algorithm~\ref{alg:slmp}. The prior policy $\pi_{\varphi}(s_t, z_t)$ takes the humanoid state $s_t$ and a latent code $z_t \in \mathbb{R}^d$ with $\|z_t\|_2=1$ and outputs a PD action. In each training iteration, we roll out the prior to obtain $s_t$, query the expert tracking controller to obtain the supervision action $a_t^*=\pi_{\text{track}}(s_t, g_t)$ using a reference goal $g_t$ drawn from the dataset, and encode the goal as
\begin{equation}
z_1 = E(g_t), \quad \|z_1\|_2 = 1. \label{eq:z1}
\end{equation}
\vspace{-5pt}
We then generate the prior action
\begin{equation}
a_{t1} = \pi_{\varphi}(s_t, z_1). \label{eq:a1}
\end{equation}
To populate the spherical latent space, we draw $\epsilon \sim \mathcal{N}(0,I)$, normalize it onto the unit sphere, and treat it as a random latent
\begin{equation}
z_2 = \frac{\epsilon}{\|\epsilon\|_2}, \quad a_{t2} = \pi_{\varphi}(s_t, z_2). \label{eq:z2}
\end{equation}
We optimize $\pi_{\varphi}$ using three losses. The imitation distillation loss matches $a_{t1}$ to the expert action:
\begin{equation}
L_{\text{distill}} = \| a_{t1} - a_t^* \|_2^2. \label{eq:distill}
\end{equation}

To provide an in-distribution signal, we introduce a discriminator $D(s_t, a_t)$ that distinguishes expert actions from non-expert actions. We train $D$ with the binary cross-entropy objective where $a_{t1}$ is treated as positive samples and $a_{t2}$ as negative samples:
\begin{equation}
L_{\text{disc}} = -\, \mathbb{E}\!\left[ \log D(s_t, a_{t1}) + \log\!\big(1 - D(s_t, a_{t2})\big) \right]. \label{eq:disc}
\end{equation}
The discriminator minimizes $L_{\text{disc}}$, while $\pi_{\varphi}$ does not optimize this adversarial loss directly.

To shape the latent manifold, we introduce a discriminator-guided local semantic consistency loss. For the pair $(z_1,z_2)$, we compute the spherical distance and neighborhood weight
\begin{equation}
d_{12} = \| z_2 - z_1 \|_2,\qquad
w_d = e^{-\beta\, d_{12}}. 
\label{eq:dist_weight}
\end{equation}
and a discriminator-based semantic weight
\begin{equation}
w_c = 1 + \big| \min\!\big(0,\, D(s_t, a_{t2})\big) \big|. \label{eq:wc}
\end{equation}
The discriminator-guided local semantic consistency loss is then defined as
\begin{equation}
L_{\text{DLSC}} = w_d \, w_c \, \| a_{t2} - a_t^* \|_2^2. \label{eq:dlscloss}
\end{equation}
The final SLMP objective minimizes distillation and semantic consistency while the discriminator is updated separately:
\begin{equation}
L_{\text{SLMP}} = \lambda_{\text{distill}} L_{\text{distill}} + \lambda_{\text{DLSC}} L_{\text{DLSC}}. \label{eq:slmp}
\end{equation}
Only $L_{\text{disc}}$ updates $D$, while $L_{\text{SLMP}}$ updates $\pi_{\varphi}$. To maintain training stability, we first train using only the $w_d$ objective until convergence, and then introduce $w_c$ for joint optimization. Note that although $\pi_{\varphi}$ does not optimize $L_{\text{disc}}$ directly, the semantic weight $w_c$ couples $\pi_{\varphi}$ with the discriminator, inducing an implicit adversarial interaction in the latent space.

% \begin{algorithm}[t]
% \caption{Spherical Latent Motion Prior (SLMP)}
% \label{alg:slmp}
% \begin{algorithmic}[1]
% \STATE Freeze expert controller $\pi_{\text{track}}$; initialize prior policy $\pi_{\varphi}$, encoder $E$, discriminator $D$
% \REPEAT
%     \STATE Sample reference goals $g_t$ from dataset; Rollout prior to obtain states $s_t$
%     \STATE Compute expert actions $a_t^* = \pi_{\text{track}}(s_t, g_t)$
%     \STATE Encode latent $z_1 = E(g_t)$ and normalize to unit sphere
%     \STATE Sample random latent $z_2 \sim \mathcal{S}^{d-1}$
%     \STATE Compute prior actions $a_{t1} = \pi_{\varphi}(s_t, z_1)$ and $a_{t2} = \pi_{\varphi}(s_t, z_2)$
%     \STATE Compute distillation loss $L_{\text{distill}}$ (Eq.~\ref{eq:distill})
%     \STATE Compute discriminator loss $L_{\text{disc}}$ (Eq.~\ref{eq:disc})
%     \STATE Compute semantic consistency loss $L_{\text{DLSC}}$ (Eq.~\ref{eq:dlscloss})
%     \STATE Update $(\pi_{\varphi}, E)$ by minimizing $L_{\text{SLMP}}$ (Eq.~\ref{eq:slmp})
%     \STATE Update $D$ by minimizing $L_{\text{disc}}$ (Eq.~\ref{eq:disc})
% \UNTIL{convergence}
% \end{algorithmic}
% \end{algorithm}

\begin{algorithm}[t]
\caption{Spherical Latent Motion Prior (SLMP)}
\label{alg:slmp}
\begin{algorithmic}[1]
\STATE Freeze expert controller $\pi_{\text{track}}$; initialize prior policy $\pi_{\varphi}$, encoder $E$, discriminator $D$
\STATE Set use\_$w_c$ $\leftarrow$ \textbf{False}

\REPEAT
    \STATE Sample reference goals $g_t$ from dataset; Rollout prior to obtain states $s_t$
    \STATE Compute expert actions $a_t^* = \pi_{\text{track}}(s_t, g_t)$
    \STATE Encode latent $z_1 = E(g_t)$ and normalize to unit sphere
    \STATE Sample random latent $z_2 \sim \mathcal{S}^{d-1}$
    \STATE Compute prior actions $a_{t1} = \pi_{\varphi}(s_t, z_1)$ and $a_{t2} = \pi_{\varphi}(s_t, z_2)$
    \STATE Compute distillation loss $L_{\text{distill}}$ (Eq.~\ref{eq:distill})

    \IF{use\_$w_c$ = \textbf{False}}
        \STATE Compute $L_{\text{DLSC}}$ without $w_c$
        \STATE Update $(\pi_{\varphi}, E)$ by minimizing $L_{\text{SLMP}}$
        \IF{converged}
            \STATE Set use\_$w_c$ $\leftarrow$ \textbf{True}
        \ENDIF
    \ELSE
        \STATE Compute semantic consistency loss $L_{\text{DLSC}}$ (Eq.~\ref{eq:dlscloss})
        \STATE Compute discriminator loss $L_{\text{disc}}$ (Eq.~\ref{eq:disc})
        \STATE Update $(\pi_{\varphi}, E)$ by minimizing $L_{\text{SLMP}}$ (Eq.~\ref{eq:slmp})
        \STATE Update $D$ by minimizing $L_{\text{disc}}$ (Eq.~\ref{eq:disc})
    \ENDIF

\UNTIL{convergence}
\end{algorithmic}
\end{algorithm}

\paragraph{Intuition.}
Pure distillation preserves expert behaviors but does not support random sampling, since most latent directions on the unit sphere lack semantic meaning and therefore cannot induce valid actions. One could instead apply a standard adversarial generator loss to increase discriminator scores on actions induced by random latents, but such adversarial feedback only provides coarse in-distribution gradients and does not offer the fine-grained supervision required to reproduce precise joint-space behaviors. We therefore employ $L_{\text{DLSC}}$ as an implicit discriminator-guided shaping objective. The geometric weight $w_d$ encourages local smoothness, while the discriminator-based weight $w_c$ increases the weight of out-of-distribution samples to impose stronger constraints. Together, these terms carve a structured latent action space on the sphere.
This structure emerges conditionally on the current state $s_t$. States with multiple plausible futures (e.g., standing) yield multi-modal latent regions, while highly constrained motions (e.g., aerial kicks) produce lower semantic variability across the sphere.

\subsection{High-Level Task: Two-Agent Combat}
\label{sec:high_level}

Similar to NCP \cite{zhu2023neural} and Smplolympics \cite{luo2024smplolympics}, SLMP is deployed as a structured low-level control prior, while a high-level policy operates in latent space and outputs latent action codes. For each agent $i \in \{1,2\}$, a high-level policy samples
\begin{equation}
    z_{t_i} \sim \pi_{h_i}(z \mid s_{t_i}), \label{eq:hlpolicy}
\end{equation}
and SLMP produces the final PD action via
\begin{equation}
    a_{t_i} = \pi_{\varphi}(s_{t_i}, z_{t_i}). \label{eq:slmp_lowlevel}
\end{equation}
This yields a clean separation of roles that the high-level policy selects latent behaviors in a compact space, while SLMP provides physics-consistent full-body execution.

The two-agent combat scenario is trained via self-play, where each agent observes its own proprioceptive state and the opponent's relative pose. The reward is strictly task-based and sparse, consisting only of rule-based hit and knockout events, with no additional reward shaping. Early termination is triggered when the agents maintain a large separation or when any agent falls, which encourages engagement under sparse rewards. Despite the minimal reward design, the high-level policies learn coordinated striking and evasion behaviors, while SLMP maintains physically consistent motor execution and transitions.
% \label{sec:high_level}
% During deployment, SLMP serves as a structured low-level control prior \cite{luo2024smplolympics}, while a high-level policy operates in latent space. For each agent $i \in \{1,2\}$, a high-level policy outputs a latent code
% \begin{equation}
%     z_{t_i} \sim \pi_{h_i}(z \mid s_{t_i}), \label{eq:hlpolicy}
% \end{equation}
% and SLMP produces the final PD action through
% \begin{equation}
%     a_{t_i} = \pi_{\varphi}(s_{t_i}, z_{t_i}). \label{eq:slmp_lowlevel}
% \end{equation}
% This enables a clean separation of roles: the high-level policy focuses on tactical decision-making in a compact latent space, while SLMP guarantees physically plausible full-body execution.

% The two-agent combat scenario is trained via self-play, where each agent observes its own proprioceptive state and the opponent's relative pose. The reward is intentionally simple: a velocity-facing component that encourages forward engagement and a sparse combat scoring term that rewards successful hits. Despite the minimal reward design, the high-level policy discovers competitive strategies, while SLMP maintains physically realistic striking, evasions, and transitions.

\section{Experiments}
\label{sec:experiments}

\begin{figure*}[htbp]
    \centering
    \includegraphics[width=1.0\linewidth]{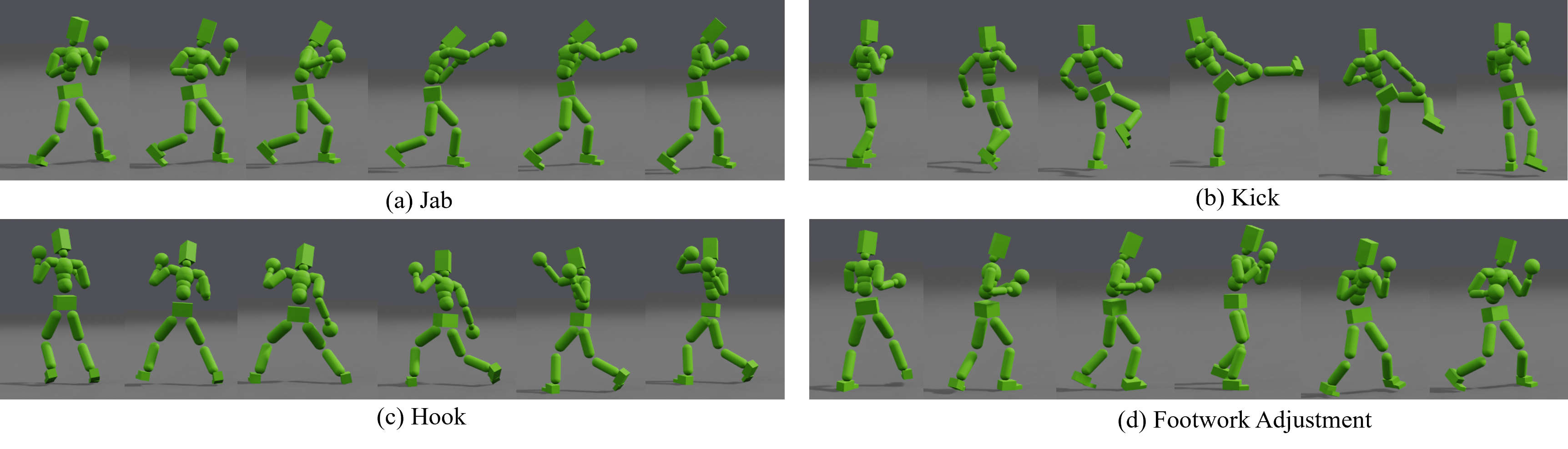}
    \caption{
Qualitative examples of random latent-conditioned rollouts generated by SLMP. 
Uniformly sampled latent codes produce diverse, stable, and physically plausible full-body motions.
Additional rollouts are provided in the supplemental video.
}
    \label{fig:qual_random_rollouts}
\end{figure*}

We base all experiments on our two-hour combat motion capture dataset introduced in Section~\ref{sec:dataset}, which provides the reference motions and tracking targets for our evaluations. We then evaluate the learned latent space in Section~\ref{sec:latent_eval}, where we quantify motion tracking performance, evaluate random rollout stability, assess semantic realism, visualize the spherical latent manifold, and perform ablations over latent representations and loss components. We demonstrate how the learned latent space supports high-level decision-making by running a two-agent combat task with simple rule-based rewards. Lastly, we show that SLMP can also be applied to real humanoid robot platforms (see Appendix~\ref{sec:real}).

\begin{figure}[htbp]
    \centering
    \includegraphics[width=1.0\linewidth]{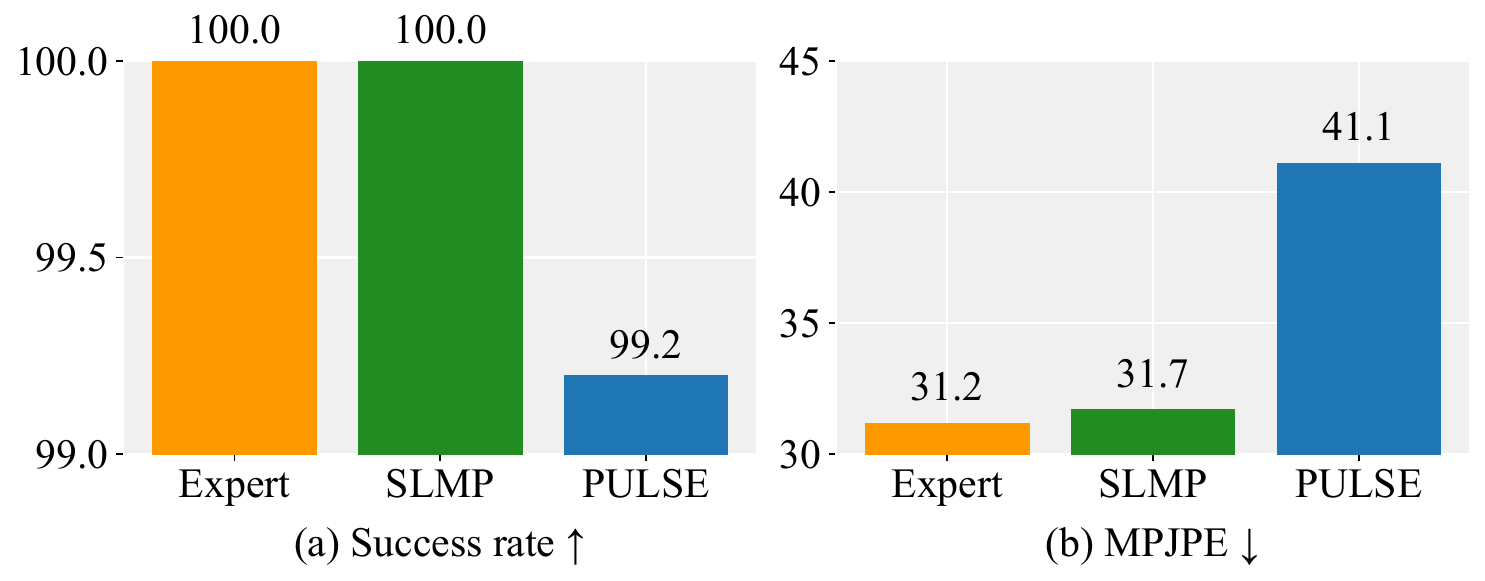}
    \caption{Latent-space motion tracking performance. We evaluate information loss by tracking reference clips through the latent space. SLMP achieves higher success and lower MPJPE than PULSE, and approaches the expert controller.}
    \label{fig:latent_tracking}
\end{figure}

% \begin{figure}[htbp]
%     \centering

%     \includegraphics[width=0.9\linewidth]{his_cover.png}

%     \caption{
%         Coverage evaluation of random latent rollouts. 
%         (a) SLMP  produces diverse behaviors that more
% evenly covers the dataset. 
%         (b) SLMP exhibits denser connections between different motions.
%     }
%     \label{fig:coverage_eval}
% \end{figure}

\subsection{Combat Motion Capture Dataset}
\label{sec:dataset}

A key contribution of this work is a new human combat motion capture dataset tailored for physics-based humanoid control. We recruited a volunteer with over three years of experience in Kickboxing. Motion data was recorded using an Xsens MVN Link inertial motion capture suit at 180 Hz \cite{xsens2013full}.

The participant was instructed to perform a comprehensive range of striking and defensive techniques. The dataset covers stances and footwork (including varying speeds and varying weight distributions), punches (jabs, crosses, hooks, uppercuts, and swings), kicks (front kicks, roundhouse kicks, side kicks, low kicks, and spinning kicks), knees, elbows, and diverse defensive maneuvers (slips, ducks, lean-backs, and checks). Crucially, to support the learning of diverse combat strategies, the actions were captured with explicit variations in speed (slow, normal, and fast/explosive) and target height (head, body, and leg). 
Additionally, we also selected a small set of boxing motions from the AMASS dataset with a total duration of about 8 minutes.
% Additionally, the dataset includes interactions with a heavy bag to capture the dynamics of physical impact, alongside solo shadow boxing sequences.
In total, we collected approximately two hours of data, which are segmented into 502 clips with durations of approximately $14$ seconds. 
% The raw motion is denoised, drift-corrected, and retargeted to our simulated humanoid model. 
For all experiments, the dataset is downsampled to 30 Hz.
% Table~\ref{tab:dataset_stats} summarizes detail statistics of the dataset.
For more details, refer to Appendix \ref{sec:a_dataset}.

% \subsection{Latent Space Evaluation}
% \label{sec:latent_eval}

% In this section we evaluate SLMP as a motion prior, independent of downstream tasks, by sampling from the latent space and rolling out motions in the physics simulator. We focus on four aspects: coverage of the combat motion dataset, semantic realism of randomly sampled motions, the structure of the spherical latent manifold, and ablations of latent parameterization and loss components.

\subsubsection{Motion Tracking Performance}

We evaluate information loss introduced by the latent space by tracking reference clips using latent codes and rolling out motions in the physics simulator. For each reference frame we compute $z_t = E(g_t)$ and reconstruct an action $a_t = \pi_{\varphi}(s_t, z_t)$, then execute it in simulation to obtain latent-space tracking. We compare against two baselines: (1) the expert tracking controller $\pi_{\text{track}}$ (upper bound), and (2) PULSE \cite{luo2024universal}, a state-of-the-art motion prior based on a VAE latent space. We measure \emph{Success} (fraction of clips tracked without falling or diverging, higher is better) and \emph{MPJPE} (mean per-joint position error between simulated and reference body markers, lower is better). Figures~\ref{fig:latent_tracking} show that SLMP yields higher Success and lower MPJPE than PULSE, approaching the expert upper bound. This indicates that SLMP introduces less information loss during latent reconstruction.

\begin{figure*}[htbp]
    \centering
    \includegraphics[width=1.0\linewidth]{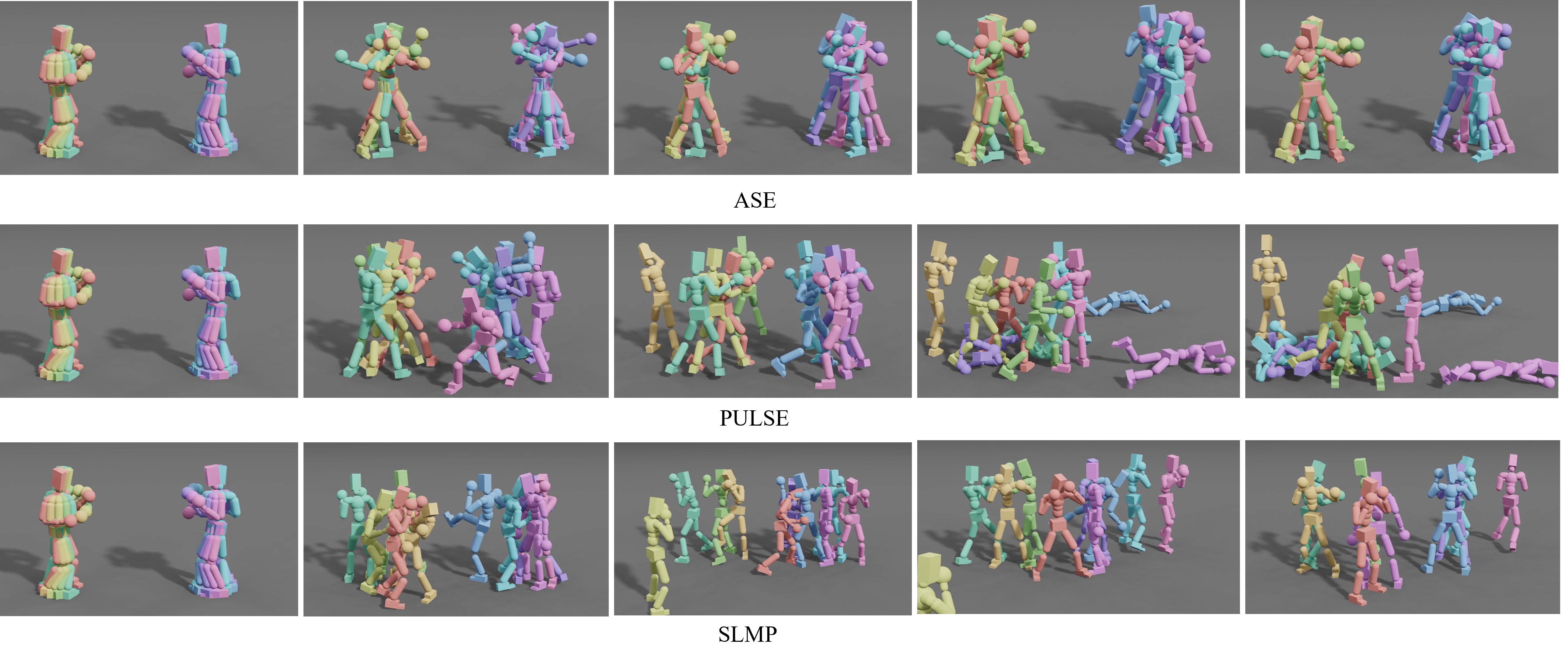}
    \caption{
        Qualitative comparison of random latent-conditioned rollouts. 
        ASE exhibits repetitive low-diversity behaviors, PULSE collapses occasionally, 
        and SLMP generates diverse and stable motions. 
        See supplemental video for full rollouts.
    }
    \label{fig:semantic_realism}
\end{figure*}

\begin{figure}[htbp]
    \centering
    \includegraphics[width=1.0\linewidth]{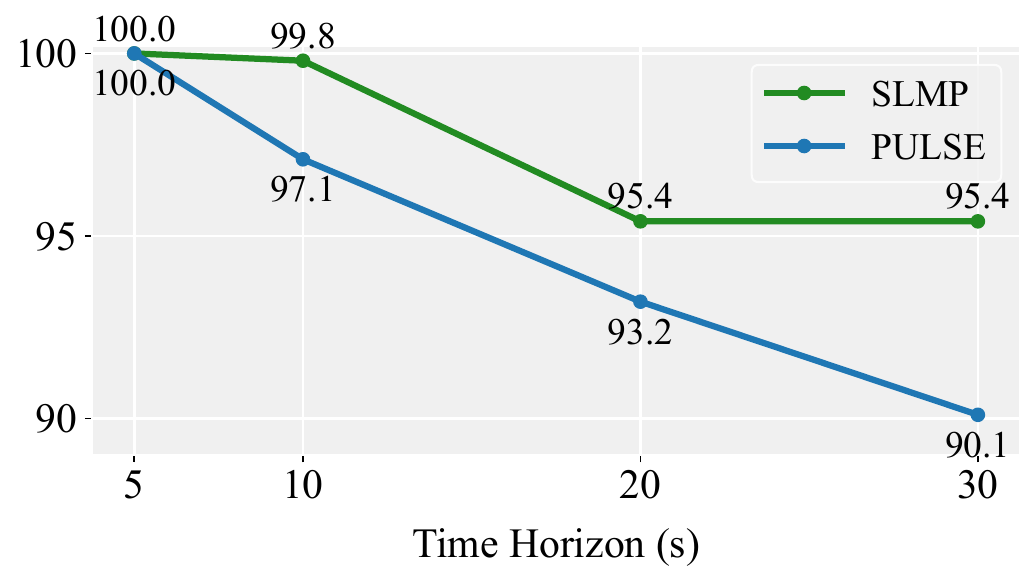}
    \caption{Random latent rollout survival curves over 1000 trials in Isaac Gym. SLMP maintains substantially higher survival rates across long horizons, whereas PULSE survival degrades due to semantic sparsity in the tails of its VAE latent space.}
    \label{fig:latent_survival}
\end{figure}

\subsection{Latent Space Evaluation}
\label{sec:latent_eval}

\subsubsection{Random Rollout Stability}
To evaluate random rollout stability, we uniformly sample latent codes $z \sim \mathcal{S}^{d-1}$ and execute the induced trajectories $a_t = \pi_{\varphi}(s_t, z)$ in Isaac Gym \cite{makoviychuk2021isaac} for fixed horizons, comparing against PULSE. Because PULSE employs a Gaussian latent distribution, the outer regions of its latent space exhibit low semantic density, and random sampling from these regions often produces invalid joint actions that rapidly lead to falls. We perform 1000 random rollouts for each method and report the fraction of trials that do not result in falls after 5s, 10s, 20s, and 30s. As shown in Figure~\ref{fig:latent_survival}, SLMP maintains substantially higher survival rates across all time horizons, whereas PULSE survival declines with rollout duration, indicating that SLMP yields a more uniformly valid latent action space for high-dynamic combat behaviors under random sampling.

% \subsubsection{Coverage Evaluation of Random Latent Rollouts}
% \label{sec:coverage_eval}

% We evaluate coverage to measure how well random samples from the latent space reproduce the diversity of combat motions in the dataset. Following ASE, we use two complementary metrics: (i) Motion Clip Coverage, which compares the distribution of clip length–conditioned motion clusters between sampled rollouts and dataset clips, and (ii) Transition Coverage, which computes a transition matrix over semantic motion clusters and measures distributional similarity. Both metrics quantify the diversity and distributional fidelity of generated motions. 
% During evaluation, we roll out each method for 500 trajectories in the physics simulator. For baselines, PULSE serves as a VAE-based latent motion prior, and ASE provides an adversarial motion prior with sphere structure. Clip-level distributions are visualized using histograms and transition matrices using heatmaps.

% Figure~\ref{fig:clip_coverage} shows that SLMP matches dataset clip statistics significantly better than PULSE and ASE, indicating broader coverage of semantic behaviors. Figure~\ref{fig:transition_coverage} reports transition matrices, where SLMP yields richer and more dataset-like transitions. These results verify that SLMP supports meaningful random sampling with high diversity.

\begin{figure}[htbp]
    \centering
    \includegraphics[width=0.9\linewidth]{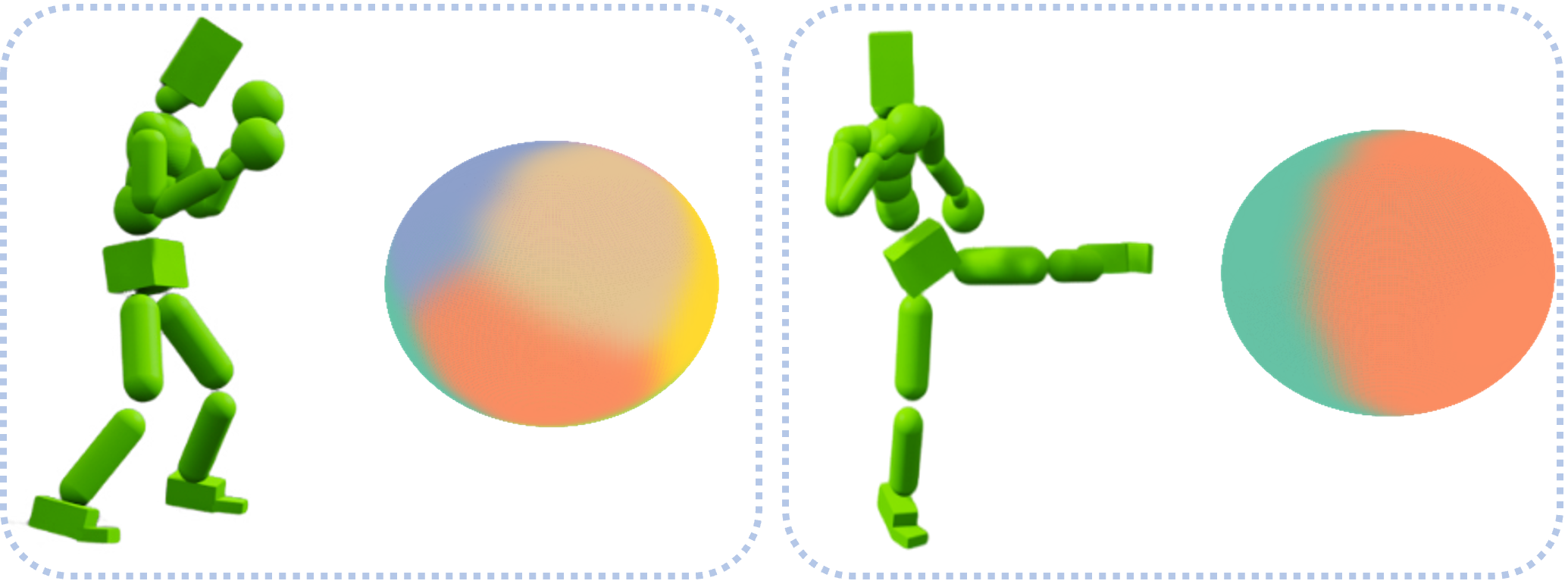}
    \caption{State-dependent spherical latent manifold.
left: a guard stance yields diverse feasible follow-ups with multiple regions on the sphere.
right: an airborne kick yields only a few feasible continuations, causing the sphere to collapse into one or two regions.}
    \label{fig:state_dependent_sphere}
\end{figure}

\subsubsection{Semantic Realism of Random Latent Rollouts}
\label{sec:semantic_realism}
We sample latent codes from the spherical prior and roll out SLMP starting from a neutral stance. Figure~\ref{fig:qual_random_rollouts} shows representative examples. SLMP generates a wide range of physically plausible and semantically interpretable combat motions, including guards, jabs, kicks, and footwork adjustments. We observe no significant artifacts such as excessive foot sliding or unrealistic joint poses.  Figure~\ref{fig:semantic_realism} illustrates qualitative results of PULSE and ASE. PULSE loses balance and collapses occasionally, while ASE remains upright but degenerates into repetitive and low-diversity motions such as small arm swings. In contrast, SLMP produces diverse and coherent motions. Refer to the supplemental video for clearer demonstrations of motion realism and diversity.

\subsubsection{Spherical latent manifold visualization}
\label{sec:sphere_vis}

We use a state-conditioned visualization to show how SLMP changes the structure of the latent sphere depending on the current pose. 
For a fixed humanoid state $s_t$, we uniformly sample latent codes $z$ on the unit sphere, pass each pair $(s_t, z)$ through the prior policy to obtain an action $a_t = \pi_{\varphi}(s_t, z)$, and cluster the resulting actions in joint space. 
We then assign a color to each cluster and paint the corresponding latent points on the sphere, which reveals how many distinct motion modes SLMP considers feasible from that state (Figure.~\ref{fig:state_dependent_sphere}).

We visualize two representative states. 
For a neutral guard stance, SLMP produces a wide variety of follow-up actions such as jabs, crosses and evasive steps, and the sphere decomposes into several colored regions with smooth boundaries. 
This pattern indicates that many qualitatively different but valid motions are available from this posture. 
In contrast, for an airborne kicking state, most latents decode to very similar landing or recovery motions, and one or two clusters cover almost the entire sphere. 
This collapse of the sphere into a small number of colors reflects the fact that the character has very few physically plausible choices while mid-air. 
Together, these visualizations show that SLMP encodes a state-dependent feasible set of actions on the latent sphere, which in turn ensures that random sampling remains stable and physically reasonable across different poses.

\vspace{-5pt}

\subsubsection{Ablation Studies}
\label{sec:ablation}

\begin{table}[htbp]
    \centering
    \caption{Ablation over latent representations. Success rate and MPJPE measure latent tracking quality. Survival measures random rollout stability under uniform sampling, defined as the percentage of 1000 randomly sampled rollouts that do not result in a fall within 10 seconds.}
    \small
    \label{tab:latent_ablation}
    % \vspace{0.5em}
    \begin{tabular}{lccc}
        \toprule
        Method & Success (\%) $\uparrow$ & MPJPE $\downarrow$ & Survival(\%) $\uparrow$ \\
        \midrule
        VAE         & 99.8 & 33.7 & 10.2 \\
        VQ-VAE      & 99.0 & 36.9 & 12.4 \\
        Sphere      & \textbf{100.0} & \textbf{31.5} & 0.1 \\
        SLMP (ours) & \textbf{100.0} & 31.7 & \textbf{99.8} \\
        \bottomrule
    \end{tabular}
\end{table}

\begin{figure}[htbp]
\centering
\includegraphics[width=1.0\linewidth]{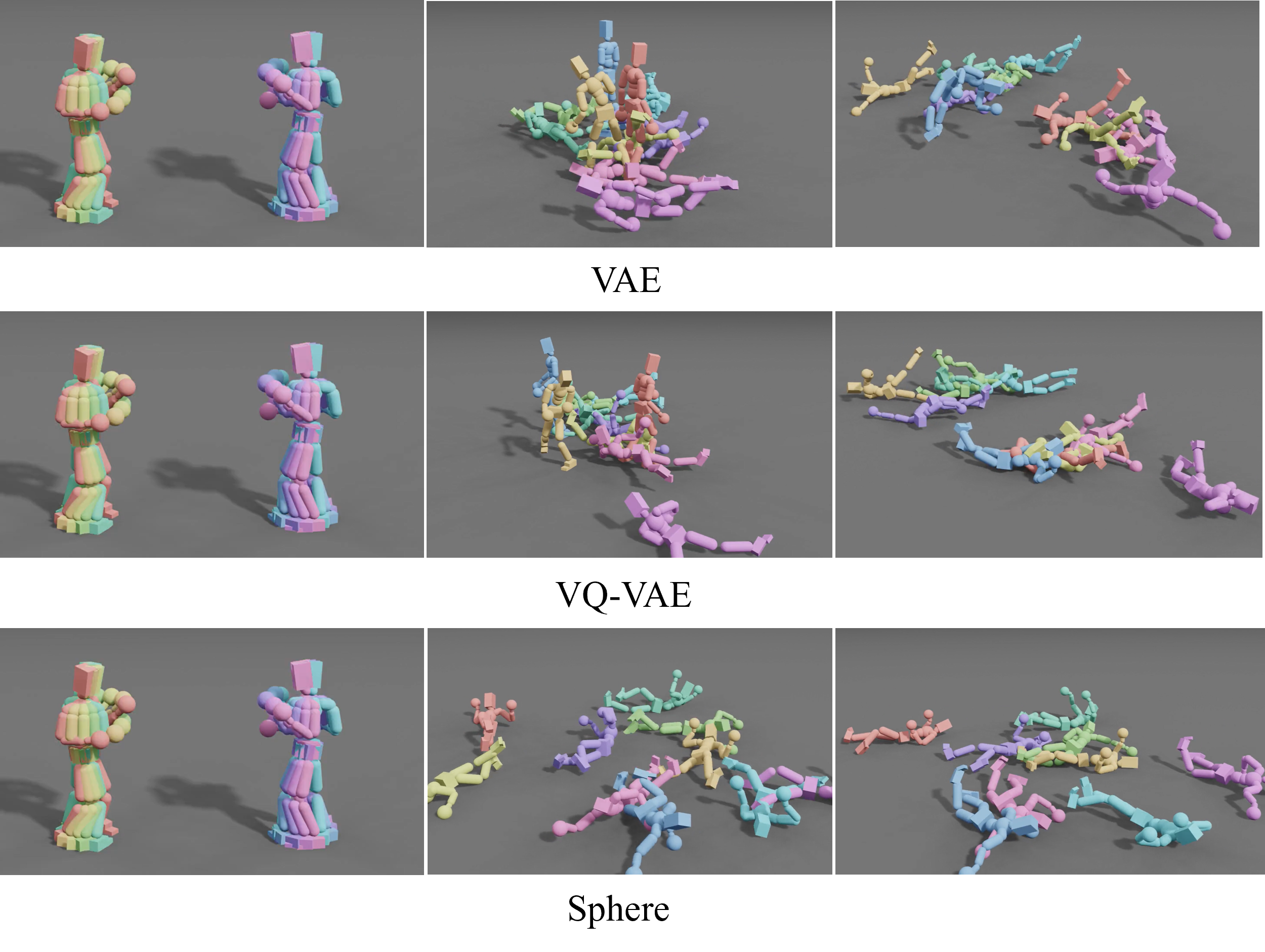}
\caption{Each row shows a rollout sampled from different latent structures.}
\label{fig:latent_representation}
\end{figure}

\paragraph{Latent Representation.}
We ablate the effect of latent representations by comparing three variants against SLMP: (i) a VAE latent, (ii) a VQ-VAE latent, and (iii) a unit-sphere latent without our proposed semantic consistency objectives. As summarized in Table~\ref{tab:latent_ablation}, VAE and VQ-VAE suffer from reconstruction-induced information loss and show unstable performance under random latent sampling. Although both methods can achieve reasonable tracking accuracy, their survival rates drop drastically, indicating that many randomly sampled latents do not correspond to physically valid or semantically meaningful motions. Imposing a spherical constraint alone removes the Gaussian prior and slightly improves tracking, but it still lacks semantic structure in the latent space. As a result, randomly sampled latents often correspond to incoherent or weakly coordinated behaviors, leading to frequent falls and near-zero survival rates. 

In contrast, SLMP maintains a structured spherical latent manifold where nearby latents correspond to semantically consistent actions. This structure preserves tracking fidelity while dramatically improving random-sampling stability, achieving near-perfect survival rates. Figure~\ref{fig:latent_representation} further visualizes qualitative rollouts. Refer to the supplemental video for clearer demonstrations.

% \paragraph{Latent Representation.}
% We ablate the effect of latent representations by comparing three variants against SLMP: (i) a VAE latent, (ii) a VQ-VAE latent, and (iii) a unit-sphere latent without our proposed semantic consistency objectives. As summarized in As shown in Table~\ref{tab:latent_ablation}, VAE and VQ-VAE suffer from reconstruction-induced information loss and show unstable performance under random latent sampling. While imposing a spherical constraint alone avoids the Gaussian prior of VAE, it still lacks semantic organization and exhibits unstable random rollouts. In contrast, SLMP achieves higher tracking fidelity and significantly improved survival under random sampling. Figure \ref{fig:latent_representation} visualizes the stability of random sampling of different latent structures. Refer to the supplemental video for clearer demonstrations.

% \begin{table}[t]
%     \centering
%     \caption{Ablation over latent representations. Success rate and MPJPE measure latent tracking quality. Survival(10s) measures random rollout stability under uniform sampling.}
%     \label{tab:latent_ablation}
%     \vspace{0.5em}
%     \begin{tabular}{lcccc}
%         \toprule
%         Method & Success (\%) & MPJPE $\downarrow$ & Survival(10s) (\%) \\
%         \midrule
%         VAE         &         &         &         \\
%         VQ-VAE      &         &         &         \\
%         Sphere      &         &         &         \\
%         SLMP (ours) &         &         &         \\
%         \bottomrule
%     \end{tabular}
% \end{table}

\begin{figure}[htbp]
    \centering
    \includegraphics[width=1.0\linewidth]{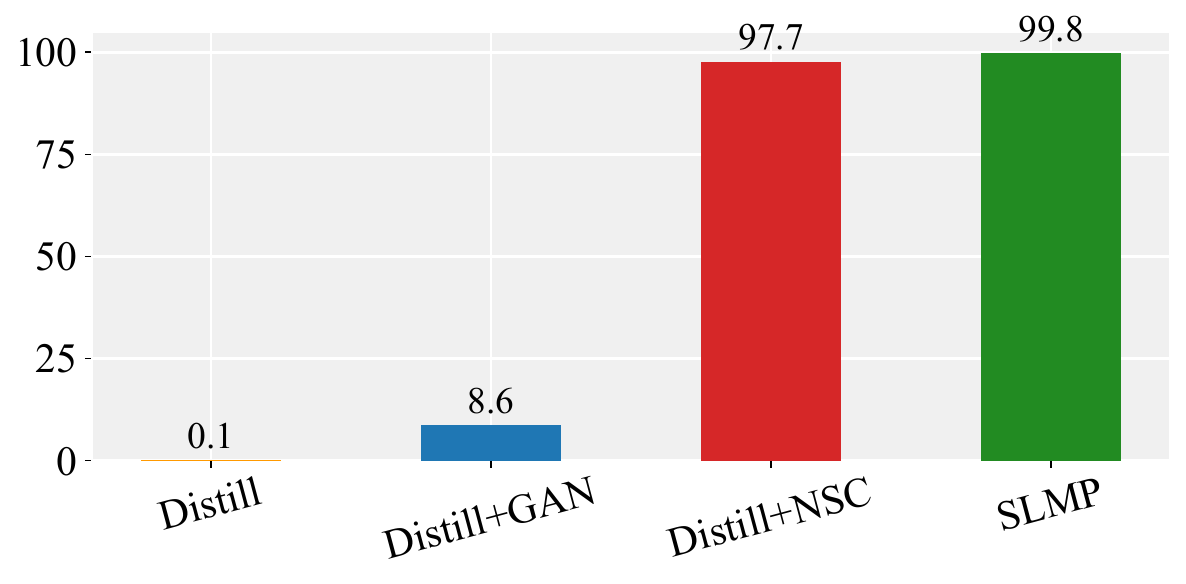}
    \caption{Survival rate at 10 seconds under random latent sampling.}
    \label{fig:loss_bar}
\end{figure}

\begin{figure}[htbp]
\centering
\includegraphics[width=1.0\linewidth]{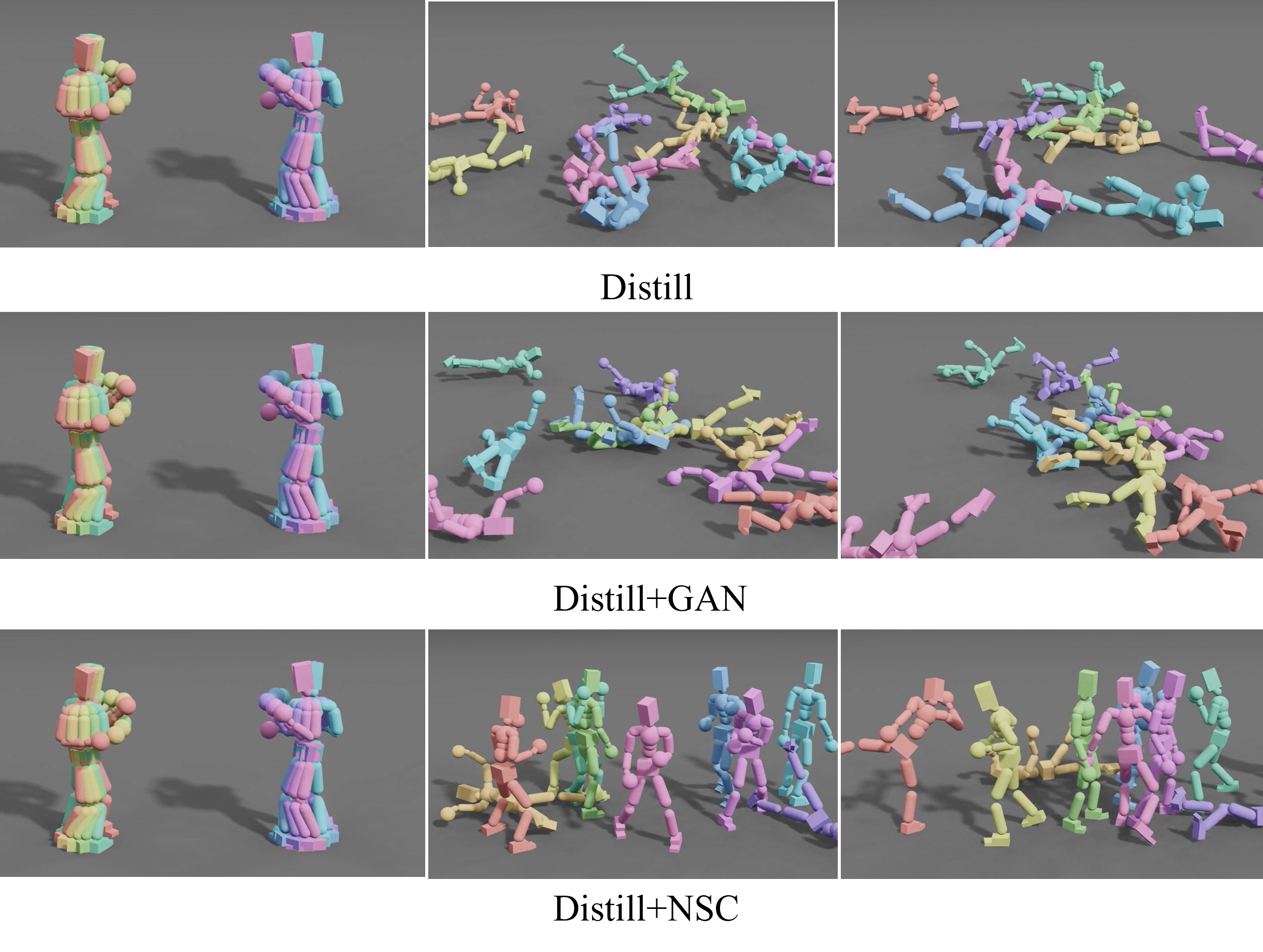}
\caption{Each row shows a rollout sampled from the latent sphere under a different training loss configuration.}
\label{fig:loss_samples}
\end{figure}

\begin{figure*}[htbp]
    \centering
    \includegraphics[width=1.0\linewidth]{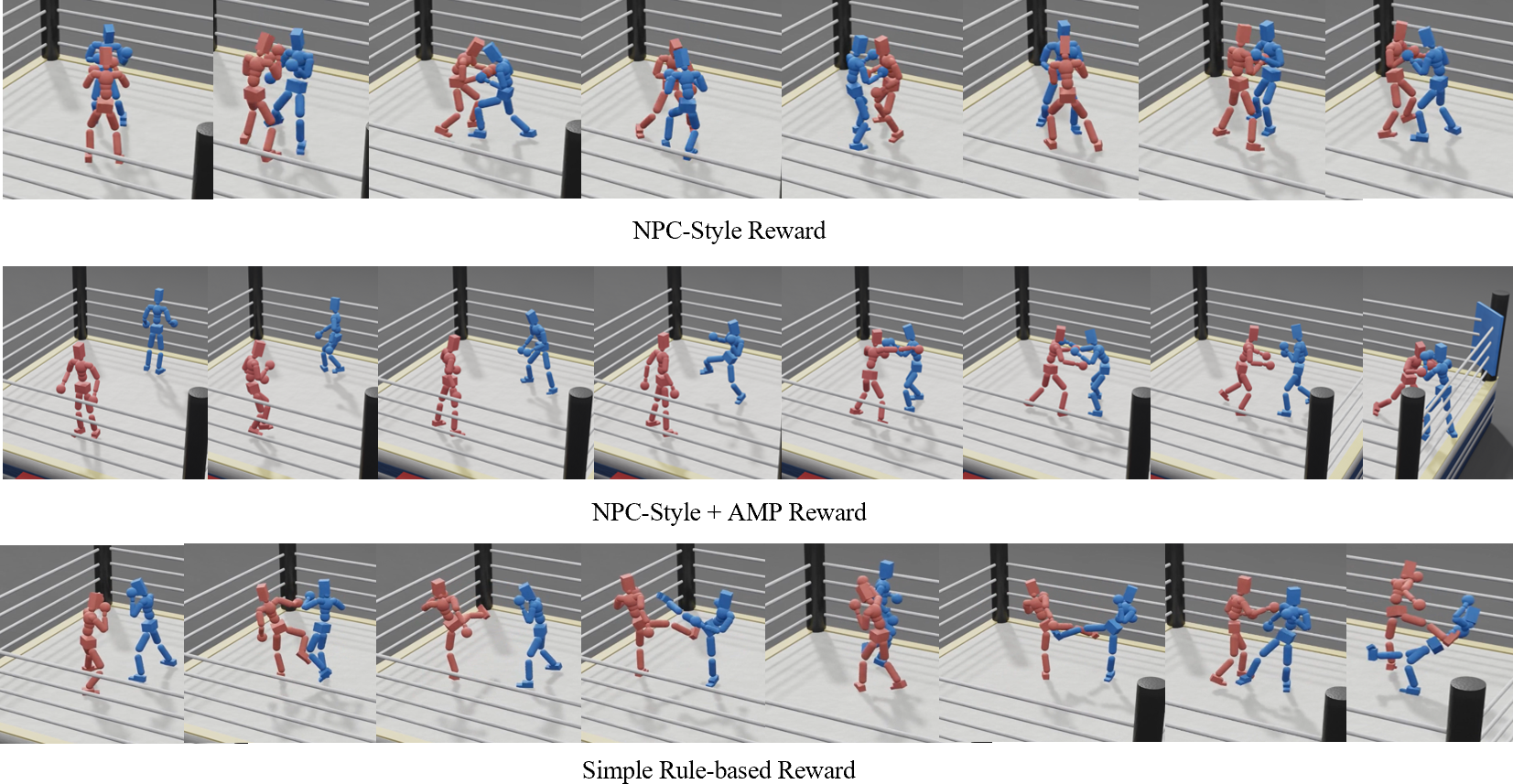}
    \caption{
    Qualitative results on the two-agent combat task using SLMP as a motion prior.
    We compare dense NCP-style rewards, Smplolympics-style rewards (NCP-style rewards augmented with AMP reward), and sparse rule-based rewards.
    NCP-style rewards lead to degenerate stationary interactions, while Smplolympics-style rewards bias the policy toward repetitive punching behaviors.
    In contrast, sparse rule-based rewards with SLMP produce diverse and human-like combat behaviors.
    }
    \label{fig:combat_sequences}
\end{figure*}

\paragraph{Loss Components.}
To isolate the contribution of each learning signal, we compare four configurations: (i) Distill, (ii) Distill+GAN, (iii) Distill+NSC (nearest semantic consistency without discriminator semantics), and (iv) SLMP. Since all four configurations achieve near-perfect latent tracking accuracy, we focus on random rollout stability and report survival rates at multiple time horizons.

We visualize the survival function at the 10-second horizon in Figure~\ref{fig:loss_bar}, where the bar plot reports the fraction of rollouts that remain fall-free. The gap highlights that SLMP maintains a substantially more stable and uniformly valid latent action space under random sampling. 

As shown in Figure~\ref{fig:loss_samples}, all three variants exhibit varying degrees of instability under random sampling. Distill lacks any regularization on the latent manifold and therefore produces unconstrained actions. Distill+GAN only receives coarse distribution-level gradients from adversarial logits, which do not provide actionable guidance for continuous control. Distill+NSC provides meaningful constraints for most samples, yet ambiguity remains when semantically different behaviors compete for neighboring regions on the sphere. SLMP achieves significantly higher survival across all horizons by combining geometric proximity with discriminator-informed semantic cues. Refer to the supplemental video for clearer demonstrations.

\subsection{High-Level Two-Agent Combat}
\label{sec:exp_combat}

We further evaluate how well the learned latent space supports high-level decision-making in a competitive setting. We compare against two reward designs commonly used in prior work. 
The first follows the combat setup in NCP~\cite{zhu2023neural}, which relies on dense shaping rewards including orientation facing, proximity, locomotion velocity, and end-effector contact.
The second adopts the reward design used in Smplolympics~\cite{luo2024smplolympics}, which augments NCP-style shaping rewards with an additional AMP-based adversarial reward to enforce motion realism.

Our setting uses only sparse rule-based combat events. A hit reward when one agent makes contact with the opponent and a knockdown reward when the opponent fall.
No imitation, adversarial, or shaping terms are used.
Despite the absence of dense shaping, SLMP enables the high-level policy to discover diverse offensive and evasive behaviors through self-play.
As shown in Figure~\ref{fig:combat_sequences}, policies trained with NCP-style rewards often converge to degenerate strategies where both agents remain in close proximity without executing coherent strikes.
Smplolympics-style rewards, while improving motion realism, bias the policy toward a narrow subset of behaviors, resulting in repetitive punching patterns with limited tactical diversity.
With SLMP and sparse rule-based rewards, agents instead exhibit structured striking, footwork, and countering behaviors.
These results demonstrate that SLMP provides a strong motion prior that reduces the need for complex reward shaping and enables expressive multi-agent behaviors.

\section{Conclusion}
We introduced the Spherical Latent Motion Prior (SLMP), a two-stage framework for learning high-fidelity and semantically structured motion priors for physics-based humanoid control. By combining expert distillation with a spherical latent manifold and a discriminator-guided semantic consistency objective, SLMP mitigates the information loss of VAE-based priors and the mode collapse tendencies of adversarial methods. Experiments on a large-scale combat motion capture dataset demonstrate that SLMP preserves fine motion detail, supports stable random sampling, and enables the emergence of human-like multi-agent combat behaviors under sparse reward signals. Beyond simulated humanoid avatars, we further validate SLMP on realistic humanoid robot models within physics simulation, demonstrating its compatibility with practical robot morphologies and control constraints. These results highlight the importance of well-structured motion priors for reducing reward engineering and simplifying high-level control in complex humanoid tasks.
Future work will explore scaling SLMP to larger and more diverse motion datasets, extending it to high-level tasks that require rich environment interaction, and pursuing real-world validation on physical humanoid robots.

\bibliography{reference}
\bibliographystyle{icml2026}

%%%%%%%%%%%%%%%%%%%%%%%%%%%%%%%%%%%%%%%%%%%%%%%%%%%%%%%%%%%%%%%%%%%%%%%%%%%%%%%
%%%%%%%%%%%%%%%%%%%%%%%%%%%%%%%%%%%%%%%%%%%%%%%%%%%%%%%%%%%%%%%%%%%%%%%%%%%%%%%
% APPENDIX
%%%%%%%%%%%%%%%%%%%%%%%%%%%%%%%%%%%%%%%%%%%%%%%%%%%%%%%%%%%%%%%%%%%%%%%%%%%%%%%
%%%%%%%%%%%%%%%%%%%%%%%%%%%%%%%%%%%%%%%%%%%%%%%%%%%%%%%%%%%%%%%%%%%%%%%%%%%%%%%
\newpage
\appendix
\onecolumn

\section{Validation on Realistic Humanoid Robot Models}
\label{sec:real}

To further evaluate the generalization ability of SLMP, 
we validate the learned motion prior on realistic humanoid robot models 
within physics simulation. 
Unlike the SMPL-based avatar used in the main experiments, 
these models reflect practical robot kinematic structures.
Specifically, we conduct experiments on the Unitree G1 \cite{unitree_g1_official}
and the ENGINEAI PM01 \cite{engineai_pm01_official} humanoid robot models.

\subsection{Random Latent Sampling}

\begin{figure*}[t]
    \centering
    \includegraphics[width=\linewidth]{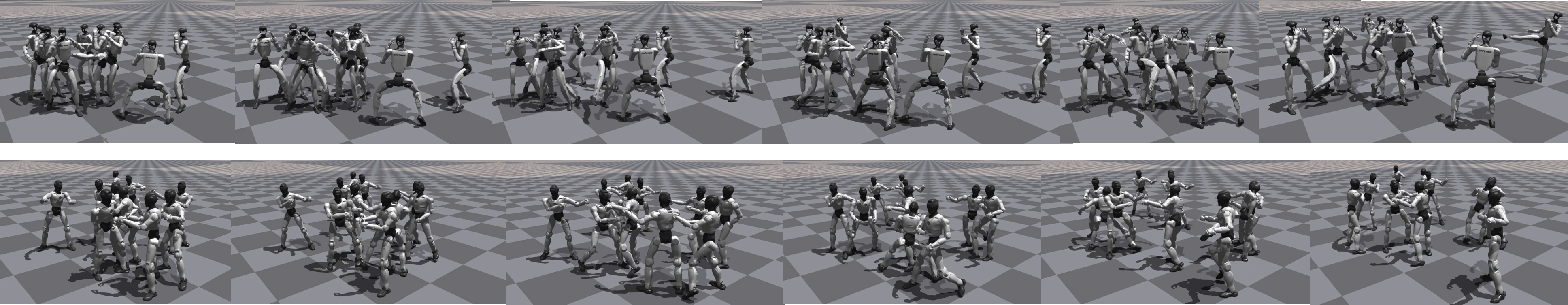}
    \caption{
    Random latent sampling on realistic humanoid robot models in simulation.
    Top: Unitree G1. 
    Bottom: ENGINEAI PM01.
    Both models can stably generate diverse motions under random latent inputs.
    }
    \label{fig:real_random}
\end{figure*}

We first evaluate random latent sampling on both robot models.
As shown in Figure~\ref{fig:real_random},
the robots are able to stably generate diverse motions under random sampling,
indicating that SLMP generalizes across different humanoid morphologies.

\subsection{Two-Agent Combat Evaluation}

\begin{figure*}[t]
    \centering
    \includegraphics[width=\linewidth]{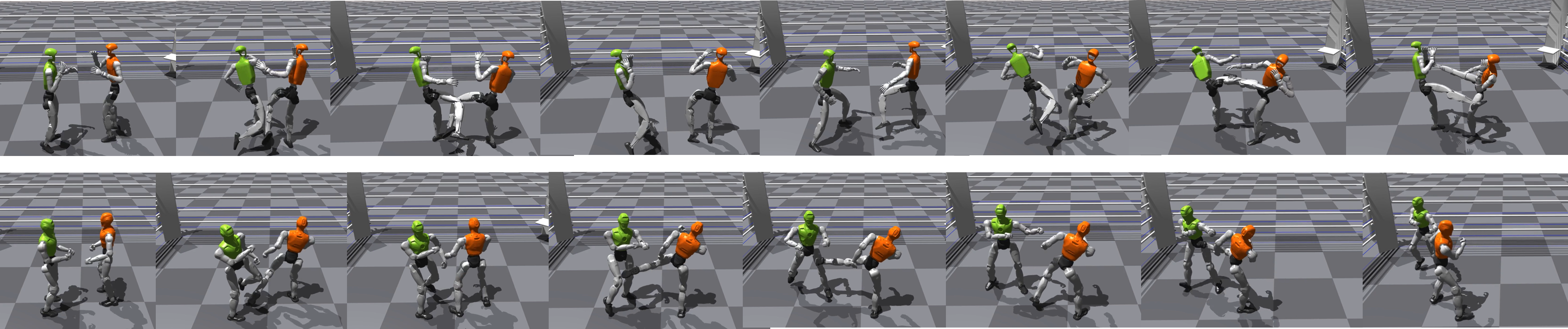}
    \caption{
    Two-agent combat evaluation on realistic humanoid robot models in simulation.
    Top: Unitree G1. 
    Bottom: ENGINEAI PM01.
    The robots generate human-like combat behaviors under the SLMP prior.
    }
    \label{fig:real_combat}
\end{figure*}

We further evaluate SLMP in a two-agent combat setting 
using the realistic robot models.
As shown in Figure~\ref{fig:real_combat},
the robots are able to generate human-like combat behaviors,
demonstrating that SLMP supports high-level interaction
across different humanoid morphologies.

Overall, these results demonstrate that SLMP is not restricted to a single 
virtual humanoid representation, but can generalize to realistic robot morphologies 
within physics simulation. 
This validation suggests strong potential for future deployment 
on physical humanoid platforms.

\section{Dataset}
\label{sec:a_dataset}

Table~\ref{tab:dataset_stats} summarizes detail statistics of the dataset.

\begin{table*}[htbp]
    \centering
    \caption{Statistics of the combat motion capture dataset (502 clips). The dataset covers diverse speeds and target levels.}
    \label{tab:dataset_stats}
    \resizebox{1.0\linewidth}{!}{%
    \begin{tabular}{lllcc}
        \toprule
        \textbf{Category} & \textbf{Action Types} & \textbf{Attributes \& Targets} & \textbf{Clips} & \textbf{Dur.} \\
        \midrule
        \textbf{Stance} & 
        Orthodox (Idle, Guard), Weight Shift & 
        Speed: Slow, Normal & 
        20 & 11s \\
        \midrule
        \textbf{Footwork} & 
        Step, Pivot ($90^\circ$), Shuffle & 
        Step Size: Small, Large & 
        62 & 11s \\
        \midrule
        \textbf{Straight Punches} & 
        Jab, Cross & 
        Target: Head, Body, Downward & 
        60 & 11s \\
        \midrule
        \textbf{Hooks \& Swings} & 
        Hook, Swing & 
        Target: Head, Body; Type: Flat/Up & 
        50 & 11s \\
        \midrule
        \textbf{Uppercuts} & 
        Uppercut (Left/Right) & 
        Target: Head, Body & 
        25 & 11s \\
        \midrule
        \textbf{Front Kicks} & 
        Front Kick (Left/Right) & 
        Target: Body; Incl. Heavy Bagss & 
        40 & 11s \\
        \midrule
        \textbf{Roundhouse} & 
        Roundhouse Kick (Left/Right) & 
        Target: Low, Mid, High; Incl. Heavy Bag & 
        55 & 11s \\
        \midrule
        \textbf{Other Kicks} & 
        Side, Low, Spinning Kick & 
        Target: Leg, Body, Head; Incl. Heavy Bag & 
        55 & 11s \\
        \midrule
        \textbf{Close Range} & 
        Knee, Elbow & 
        Target: Body, Head; Direction: Multi & 
        30 & 11s \\
        \midrule
        \textbf{Defense} & 
        Guard, Slip, Duck, Lean-back, Check & 
        Intensity: Low to High & 
        70 & 11s \\
        \midrule
        \textbf{Combinations} & 
        Jab-Cross-Hook variants, Counters & 
        Seq: 2-5 strikes; Mixed Defense & 
        35 & 10-20s \\
        \midrule
        \textbf{Total} & 
        \textbf{All Categories} & 
        \textbf{--} & 
        \textbf{502} & \textbf{$\sim$2.0h} \\
        \bottomrule
    \end{tabular}%
    }
\end{table*}

\section{Implementation Details}

\subsection{Network Architectures}
All networks are implemented as multilayer perceptrons (MLPs) with component-specific depth and width, as summarized in Table~\ref{tab:network_arch}. The two high-level combat controllers $\pi_{h_1}$ and $\pi_{h_2}$ share the same architecture.

\subsection{Training Hyperparameters}
We use PPO to train the expert tracking controller $\pi_{\text{track}}$. The latent prior is trained using a DAgger-style \cite{ross2011reduction} online imitation procedure, where the student policy rolls out trajectories and the expert provides on-policy supervision.
 Table~\ref{tab:hyperparams} summarizes key hyperparameters.

\begin{table}[htbp]
\centering
\caption{Network architectures used in our framework. All networks are implemented as multilayer perceptrons (MLPs).}
\label{tab:network_arch}
\vspace{6pt}
\begin{tabular}{l l l}
\toprule
\textbf{Component} & \textbf{Hidden Units} & \textbf{Activation} \\
\midrule
$\pi_{\text{track}}$ 
& [2048, 1536, 1024, 1024, 512, 512] 
& SiLU \\
$E(\cdot)$ 
& [512, 256] 
& ReLU \\
$\pi_{\varphi}$ 
& [4096, 2048, 1024, 1024, 512, 512] 
& SiLU \\
$D(\cdot)$ 
& [4096, 2048, 1024, 512] 
& ReLU \\
$\pi_{h_1}, \pi_{h_2}$ 
& [2048, 1024, 512] 
& SiLU \\
\bottomrule
\end{tabular}
\end{table}

\begin{table}[htbp]
\centering
\caption{Key hyperparameters used in our experiments.}
\label{tab:hyperparams}
\vspace{6pt}

\begin{tabular}{lccccc}
\toprule
\multicolumn{6}{c}{\textbf{PPO Hyperparameters}} \\
\midrule
Batch Size & Env Num & LR & $\gamma$ & Clip $\epsilon$ \\
$32768$ & $1024$ & $5\times10^{-5}$ & $0.99$ & $0.2$ \\
\midrule

\multicolumn{6}{c}{\textbf{SLMP Hyperparameters}} \\
\midrule
$\lambda_{\text{distill}}$ & $\lambda_{\text{disc}}$ & Disc LR & $\lambda_{\text{DLSC}}$ & $\beta$ & $d$ \\
$1.0$ & $1\times10^{-4}$ & $5\times10^{-5}$ & $1.0$ & $0.1$ & $64$ \\
\bottomrule
\end{tabular}

\end{table}

\subsection{Software and Hardware Setup}
All simulations are implemented in Isaac Gym with a physics timestep of 60 Hz, and PD control runs at the same frequency. Motion data is downsampled to 30 Hz for training. Experiments are conducted on a machine with Ubuntu 22.04, Python 3.8, and PyTorch 2.1. Training is performed on a server equipped with two NVIDIA RTX 4090 GPUs. Training the tracking controller typically takes 24 hours, training the latent prior takes 12 hours, and training the two-agent combat policy requires approximately 8 hours.

\section{Detailed Task Formulation}

We provide additional implementation details for the three training stages. We focus on state/observation design and reward functions, which are most critical for reproducibility.

% =====================================================
\subsection{Stage 1: Motion Tracking Controller}
\label{app:stage1_tracking}

We train the expert motion tracking controller following PHC \cite{luo2023perpetual} under a goal-conditioned RL formulation. The task is an MDP where the policy outputs PD targets for each actuated DoF, and the simulator applies torques via a PD controller.

\paragraph{State.}
The simulation state is defined as
\begin{equation}
s_t \triangleq (s^p_t, s^g_t),
\end{equation}
where $s^p_t$ is the humanoid proprioception and $s^g_t$ is a reference-driven goal. The proprioceptive state is
\begin{equation}
s^p_t \triangleq (q_t, \dot{q}_t),
\end{equation}
where $q_t$ and $\dot{q}_t$ denote the humanoid pose and velocity in simulation.

The goal state uses the discrepancy between the next-step reference quantities and the simulated counterpart:
\begin{equation}
s^{g\text{-rot}}_t \triangleq
(\hat{\theta}_{t+1}\ominus \theta_t,\;
 \hat{p}_{t+1}-p_t,\;
 \hat{v}_{t+1}-v_t,\;
 \hat{\omega}_{t+1}-\omega_t,\;
 \hat{\theta}_{t+1},\;
 \hat{p}_{t+1}),
\end{equation}
where $\ominus$ computes the rotation difference.

All quantities in $(s^p_t, s^g_t)$ are normalized w.r.t. the humanoid's current facing direction and root position, as in PHC \cite{luo2023perpetual}.

\paragraph{Reward.}
The per-step reward is composed of a task imitation term, an AMP-style discriminator term, and an energy penalty:
\begin{equation}
r_t = 0.5\, r^g_t + 0.5\, r^{\mathrm{amp}}_t + r^{\mathrm{energy}}_t.
\end{equation}
For motion tracking, the task reward takes the form
\begin{equation}
\begin{aligned}
r^{g\text{-imitation}}_t
&= w_{\mathrm{jp}}\exp\!\big(-100\|\hat{p}_t - p_t\|\big)
 + w_{\mathrm{jr}}\exp\!\big(-10\|\hat{q}_t \ominus q_t\|\big) \\
&\quad + w_{\mathrm{jv}}\exp\!\big(-0.1\|\hat{v}_t - v_t\|\big)
 + w_{\mathrm{j}\omega}\exp\!\big(-0.1\|\hat{\omega}_t - \omega_t\|\big),
\end{aligned}
\end{equation}
which measures discrepancies of link translations, rotations, linear velocities, and angular velocities between the simulated motion and the reference.

The energy penalty is defined as
\begin{equation}
r^{\mathrm{energy}}_t
= -0.0005 \sum_{j\in \text{joints}} \|\mu_j \omega_j\|^2,
\end{equation}
where $\mu_j$ and $\omega_j$ are the joint torque and joint angular velocity, respectively.
The style reward $r^{\mathrm{amp}}_t$ is computed using an AMP-style discriminator with the same observation design and training objective as PHC \cite{luo2023perpetual}.

% =====================================================
\subsection{Stage 2: SLMP Latent Prior}

\paragraph{State and Goal.}
The policy input uses the same proprioceptive state definition as Stage 1. The goal $g_t$ is derived from the reference motion and encoded by the goal encoder as described in the main paper. No additional observations are introduced in this stage.

% =====================================================
\subsection{Stage 3: Two-Agent Combat}

\paragraph{Observation.}
The observation design largely follows the boxing setup in SMPLOlympics~\cite{luo2024smplolympics}. 
Each agent receives both self and opponent information in an egocentric frame, including:

\begin{itemize}
    \item Self proprioception: joint poses, velocities, and root state,
    \item Opponent root pose and velocity relative to the agent,
    \item Relative positions between the agent’s striking limbs (hands and feet) and the opponent’s scoring regions (head and torso),
    \item Contact force magnitudes on key body parts.
\end{itemize}

All quantities are expressed in a heading-aligned local frame.

\paragraph{Reward Function.}
We use simple rule-based sparse rewards focused on effective striking.

\textbf{Hit reward.}
A positive reward is given when an agent’s hand or foot comes within $0.3$\,m of the opponent’s head or torso and the corresponding contact force exceeds $30$. 
The reward magnitude is proportional to the measured contact force.

\textbf{Hit penalty.}
A symmetric penalty is applied when the agent’s own head or torso is hit under the same conditions.

\textbf{Knockdown reward.}
A bonus of $+50$ is given when the opponent falls. 
Conversely, a penalty of $-50$ is applied when the agent falls.

\paragraph{Termination Conditions.}
Episodes terminate under the following conditions:

\begin{itemize}
    \item Knockdown of either agent,
    \item pelvis-to-pelvis distance below $0.3$\,m for more than $1$\,s,
    \item Distance between an attacking limb and the opponent’s scoring regions below $0.3$\,m for more than $1$\,s (to prevent reward farming),
    \item During early training, if the pelvis distance exceeds $1.2$\,m.
\end{itemize}

\paragraph{Self-Play Setup.}
We adopt an alternating self-play scheme with two separate policy instances. 
At any time, one policy is designated as the learning agent while the other serves as a fixed opponent. 
During training, the learning agent is updated using PPO while the opponent policy remains frozen. 
Every 250 epochs, the roles are swapped, and the previously fixed policy becomes trainable while the other is held fixed.

This alternating update stabilizes competitive learning by preventing both agents from simultaneously drifting, and encourages continual adaptation against a progressively improving opponent. 
Both policies are initialized identically but evolve independently during training.

\end{document}